\documentclass[a4paper,fleqn]{cas-dc}
 \usepackage[square, comma, sort&compress, numbers]{natbib}
\usepackage{epstopdf}
\usepackage{color}
\usepackage{algorithmic}
\usepackage{algorithm}
\usepackage{booktabs}

\def\tsc#1{\csdef{#1}{\textsc{\lowercase{#1}}\xspace}}
\tsc{WGM}
\tsc{QE}
\tsc{EP}
\tsc{PMS}
\tsc{BEC}
\tsc{DE}
\begin{document}
\let\WriteBookmarks\relax
\def\floatpagepagefraction{1}
\def\textpagefraction{.001}
\shorttitle{}
%\shortauthors{Author et~al.} %% 缩略作者 自己名字， 比如： 张三 = S. Zhang

%% 标题
\title [mode = title]{Novel total hip surgery robotic system based on self-localization and optical measurement}                      
%%\tnotemark[1,2]

%%\tnotetext[1]{This document is the results of the research project funded by the National Science Foundation.}

%%\tnotetext[2]{The second title footnote which is a longer text matter to fill through the whole text width and overflow into another line in the footnotes area of the first page.}

%% 作者顺序
%% 1
%\author[1]{\textcolor[RGB]{0,0,1}{Knowledge can't get into the brain}}
%\fnmark[1] %%第几作者
%\credit{}%%本文的贡献
%\address[1]{Department}  

%% 2
%\author[1,2]{\textcolor[RGB]{0,0,1}{name1}}
%\fnmark[2]
%\cormark[1]%%通讯作者星标
%\ead{123456@gmail.com}
%\address[1]{Department address}

%%3
%\author[1]{\textcolor[RGB]{0,0,1}{name1}}
%\fnmark[1]

%% 4
%\author[2]{\textcolor[RGB]{0,0,1}{name2}}
%\fnmark[2]
%\address[2]{Department address}
\author[author1]{Weibo Ning}
\author[author1]{Jiaqi Zhu}
\author[author2]{Hongjiang Chen}
\author[author1]{Weijun Zhou}
\author[author1]{Shuxing He}
\author[author1]{Yecheng Tan}
\author[author1]{Qianrui Xu}
\author[author1]{Ye Yuan}
\author[author2]{Jun Hu\corref{cor1}}
\author[author1]{Zhun Fan\corref{cor2}}
\address[author1]{Shantou University,Shantou,China}

\address[author2]{The First Affiliated Hospital of Medical College of Shantou University,Shantou,China}

\cortext[cor1]{Corresponding author}
\cortext[cor2]{Corresponding author}

\cortext[cor1]{Corresponding author:} %% 首页左下角通讯作者
%%\cortext[cor2]{Principal corresponding author} 

%%\fntext[fn1]{This is the first author footnote. but is common to thirdauthor as well.}
%%\fntext[fn2]{Another author footnote, this is a very long footnote and it should be a really long footnote. But this footnote is not yet sufficiently long enough to make two lines of footnote text.}

%%\nonumnote{This note has no numbers. In this work we demonstrate $a_b$ the formation Y\_1 of a new type of polariton on the interface between a cuprous oxide slab and a polystyrene micro-sphere placed on the slab.}

%%摘要
\begin{abstract}
This paper presents the development and experimental evaluation of a surgical robotic system for total hip arthroplasty (THA). Although existing robotic systems used in joint replacement surgery have achieved some progresses, the robot arm must be situated accurately at the target position during operation, which depends significantly on the experience of the surgeon. In addition, handheld acetabulum reamers typically exhibit uneven strength and grinding file. Moreover, the lack of techniques to real-time measure femoral neck length may lead to poor outcomes. To tackle these challenges, we propose a real-time traceable optical positioning strategy to reduce unnecessary manual adjustments to the robotic arm during surgery, an end-effector system to stabilise grinding, and an optical probe to provide real-time measurement of the femoral neck length and other parameters used to choose the proper prosthesis. The lengths of the lower limbs are measured as the prosthesis is installed. The experimental evaluation results show that, based on its accuracy, execution ability, and robustness, the proposed surgical robotic system is feasible for THA.
\end{abstract}

%\begin{graphicalabstract}
%%\includegraphics{figs/grabs.pdf} %%图片摘要地址路径
%\end{graphicalabstract}

%%高亮
%begin{highlights}
%\item highlights 1.
%\item highlights 2.
%\item highlights 3.
%\end{highlights}

%% 关键词

% 此指令为生成标题格式，不可删除
\maketitle  

%% 1.引言
\section{Introduction}

%%\par{文本内容}换行并缩进
\par{ Total hip arthroplasty (THA) is one of the most important methods used to alleviate the symptoms of patients with hip diseases \cite{petis2015surgical, cao2016review, hall2022day}. Unfortunately, the quality of traditional THA surgery relies heavily on the experience of surgeons. Because its intraoperative field of vision is limited, it can easily lead to incorrect size choices and placement of prostheses. There may also be other complications, such as early dislocation and postoperative leg length discrepancy, among others \cite{kumar2021does, marcovigi2022dislocation, parratte2007validation}. To address these problems, Lewinnek et al.~\cite{lewinnek1978dislocations} suggested reducing postoperative complications by focusing upon the normal angle range of acetabular prosthesis placement, i.e. the abduction angle is 40°±10°, whereas the anteversion angle is 15°± 10°. During the operation, it is difficult for a surgeon to correctly place the prosthesis because of the lack of positioning information \cite{kalteis2005greater, saxler2004accuracy}. Thus, Widmer et al.~\cite{widmer2004simplified} used a trigonometric function relationship to determine the positioning of an acetabular cup using X-ray imaging. However, the disadvantage is radiation exposure and additional skin incisions, which increases other accidental risks. On the other hand, Dechenne et al.~\cite{dechenne2005novel} used a mechanical equipment to select anatomical bone markers to correctly place an acetabular prosthesis. However, the device is based on an ideal surgical scheme and hard to generalize. \\
	\indent Postoperative leg length discrepancy remains the most common cause of patient dissatisfaction and malpractice litigation in hip arthroplasty \cite{stewart2022comparison, plaass2011influence, asayama2005reconstructed}. It is important to choose appropriate lengths for femoral head prostheses and to minimise inequalities between both lower limbs after the operation, enabling patients to recover faster. Several methods have been proposed for adjusting the length of the femoral neck during surgery. For example, Kyoichio et al.~\cite{ogawa2014accurate} used a measuring device similar to a Vernier caliper. Tagomori et al. ~\cite{tagomori2019new} used a ruler to measure the marked position of the posterior acetabulum prior to dislocation. However, the disadvantage of these mechanical structures is that the deployment of measuring equipment requires direct intervention on the surgical lesion of the patient, which may interfere with the operation performed by the surgeon. In addition, these devices are not convenient for surgeons to read.\\
	\indent Recently, the clinical efficacy of surgical robots has become a hot topic joint surgery \cite{xia2021}. It has been reported that robot-assisted THA can improve the precision of prosthesis placement and have advantages over conventional THA in reducing the difference between the lower limbs \cite{xu2022s, sicat2022intraoperative, davies2006active}. A wider implementation of robotic surgery is thus expected as surgeons become more familiar with this technology \cite{banerjee2016robot, li2022hurwa}. 
The clinical outcome may be improved through the selection of an appropriate grinding angle and femoral neck length, thus reducing inequality in the lengths of the lower limbs after the operation. Currently, commercial hip surgery robots use visual navigation technologies to guide surgeons to accurately and quickly place prostheses with the help of external tracking devices \cite{rowan2018prevention, sugano2013computer}. In early 2007, RoboDoc (Integrated Surgical Systems, USA) obtained results comparable to those of traditional surgical procedures \cite{schulz2007results}. Hip surgery robots currently available in the market, such as MAKO (Stryker, USA), have also achieved satisfactory results in assisting THA. Their navigation systems ensure accurate placement of acetabular prostheses \cite{subramanian2019review, hadley2020robotic}. However, they still have the following three shortcomings: First, the cost is high, and thus the popularity is low. Second, during the operation, surgeons must drag the robot arm into position, which requires good clinical experience. Third, because the length of the femoral neck and the difference between both lower
limbs cannot be measured during the operation, the manipulation process is highly dependent on the registration process for image space and surgical space. As a result, the operation accuracy also depends on the registration accuracy. Moreover, the real-time registration process is complicated, and the learning curve for the surgical procedure is long \cite{kouyoumdjian2020current, perets2020current}.\\
	\indent Therefore, in this paper, based on the principle of optical localisation, a method based on real-time tracking and positioning is proposed to facilitate accurately measuring distances and operating robotic arms during surgery, which can not only improve the efficiency of prosthesis positioning, but also significantly reduce the chance of inequality in the lengths of the lower limbs after the operation. In summary, an imageless surgical navigation system based on optical positioning is proposed in this work, in which the femoral neck length is measured in real time using a probe, so that postoperative complications can be largely reduced via comparisons between the lengths of the lower limbs during the surgery. In addition, an end-effector system is developed to enable stable grinding and improve surgical safety.}

\section{FRAMEWORK DESIGN}
\subsection{Total Hip Arthroplasty Robot System}
In this study, we developed a surgical robot based on optical positioning and a multi-degree-of-freedom flexible robotic arm to help surgeons eliminate subjective errors and improve the quality of hip-replacement surgery. The framework of the robotic system consists of preoperative planning and intraoperative execution (Figure 1). After preoperative 3D CT reconstruction, grinding path planning and hand-eye calibration were performed. The calibration and navigation systems were designed based on an optical positioning system to be used during the operation. It is very important to obtain accurate target position information from the calibrator and intraoperative navigation system to reduce errors during surgery. To facilitate the surgery operations, an end-effector system to be applied on the 6-DOF flexible robotic arm was also designed and developed, which had robust signal acquisition and feedback functions. Further details are provided in sections II-B, II-C, and II-D.

%% 图1  后面要加就自己复制这个改改
\begin{figure*}
	\centering
	\includegraphics[width=17.6cm]{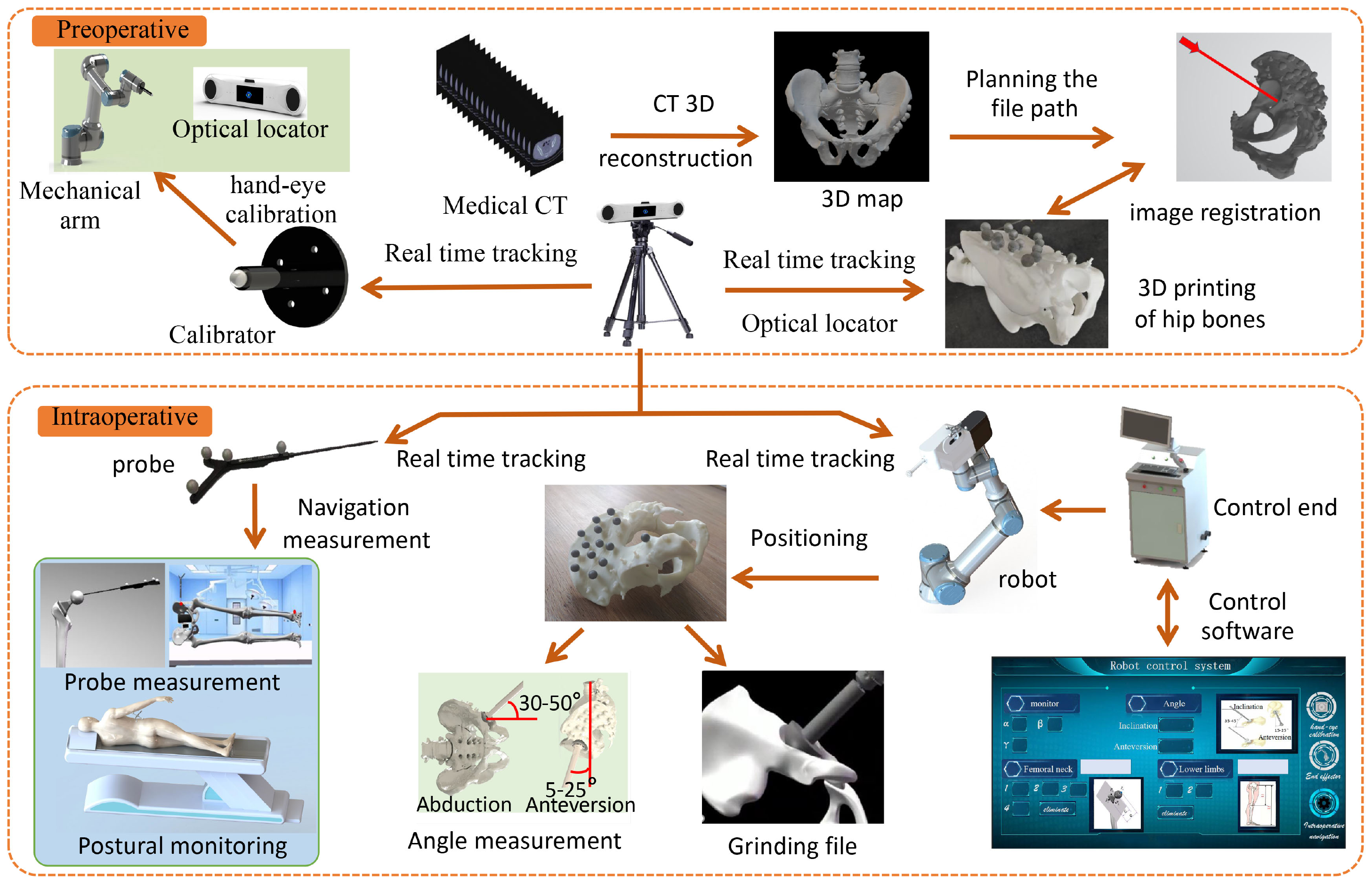}
	\caption{Frame of autonomous-navigation THA robot system.}
	\label{fig:1}
\end{figure*}
\subsection{Calibrators and Image Processing}
The calibration tool developed in this study is fixed onto the end of the robotic arm through a flange (Figure 2). A new coordinate system was obtained through corresponding adjustments on the offset of the centre of the end flange. Therefore, the dimension parameters from the end flange connection centre to the marker ball centre would be calculated using the design drawing of the calibrator. The coordinate information of the marker ball centre was read from the base coordinates of the manipulator. The movement of the robotic arm was controlled with a preset step length, and the sequence of position information of the centre of the marker ball installed in the optical positioning instrument with reference to the robotic arm base coordinate system was then recorded. Based on singular value decomposition, as demonstrated in literature \cite{sorkine2017least, ho2013finding}, a novel algorithm with added error filtering(as illustrated in Algorithm 1) was used to calculate the hand-eye conversion relationship $ H_{base}^{cam}$. In accordance with the transformation relationship of the coordinates $H_{base}^{pic}=H_{cam}^{pic} \cdot H_{base}^{cam}$, the transform matrix  $ H_{base}^{pic}$ was calculated to complete the path positioning. Furthermore, after the surgeon planned a proper grinding path on the image, the robot could move accurately according to it. Particularly, $ H_{base}^{pic}$   is the transformation from the image space to the base coordinate system of the robot arm.   $ H_{cam}^{pic}$ is the transformation from the image space to the coordinate system of the optical locator.  $ H_{base}^{cam}$ is the transformation from the optical locator to the base coordinate system of the robot arm.

\begin{figure}
	\centering
	\includegraphics[width=8.2cm]{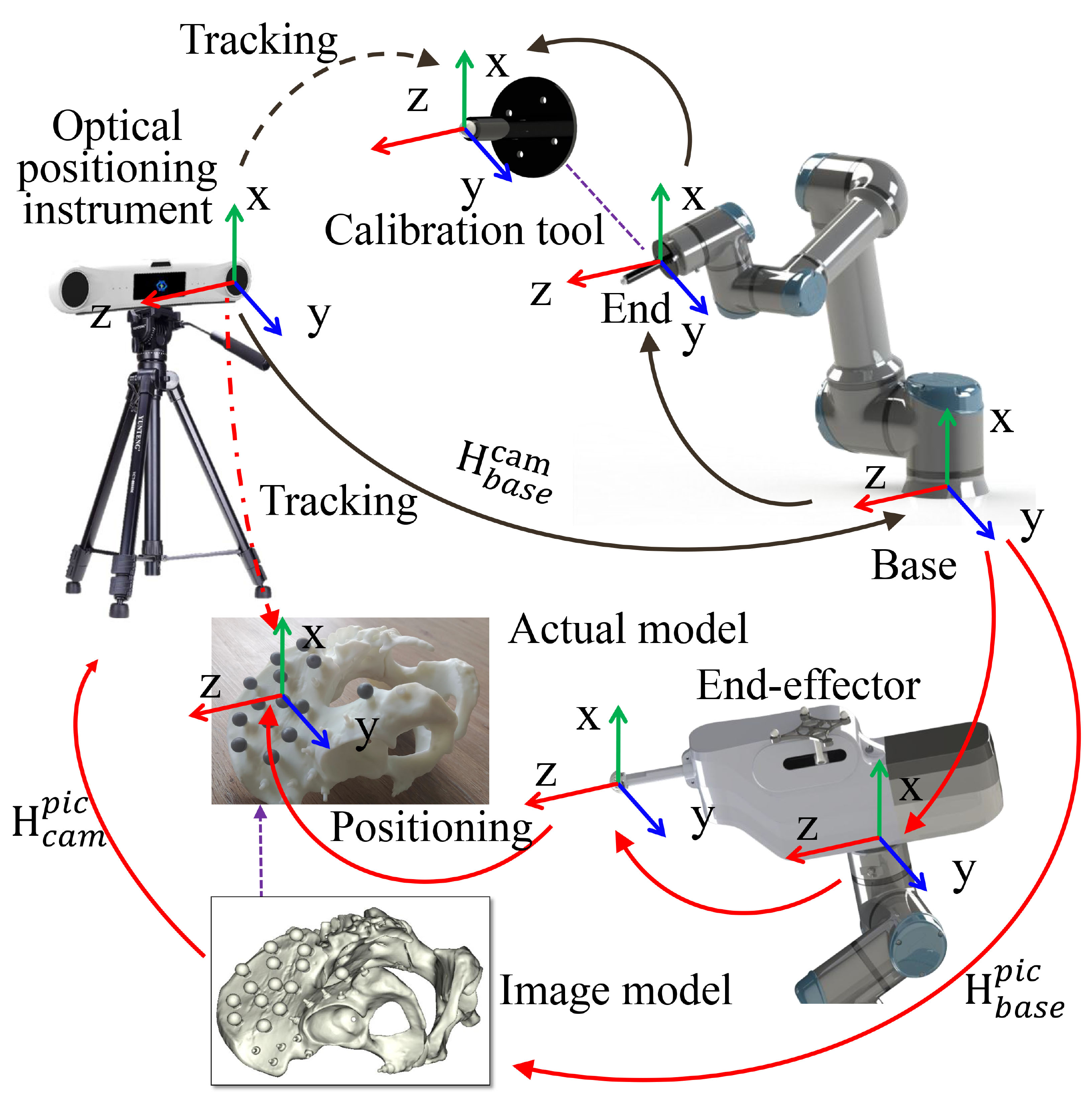}
	\caption{Robot arm - Optical locator - Calibration tool - Calibration of end-effector. Setup for coordinate system transformation of robot for target positioning.}
	\label{fig:2}
\end{figure}

\begin{algorithm}
	\renewcommand{\algorithmicrequire}{\textbf{Require:}}
	\caption{Robotic calibration process for total hip surgery}
	\label{alg:1}
	\begin{algorithmic}[1]
		\REQUIRE  Plan the initial point and step size of the robotic arm movement
		%	\ENSURE $U^{p}$, $V^{p}$, $b^{p}$
		\STATE The robot arm moves according to the corresponding step length and records the hand-eye point pair set $P$,~$Q$
		\STATE $P$ and $Q$ are averaged: $ \bar P$ , $ \bar Q$
		\STATE Subtract the mean (center point) from the data:$\widetilde{P}= P-\bar P , \widetilde{Q}= Q-\bar Q$
		\STATE  Matrix calculation: $H= \mathbf{\widetilde{P}}^\top\cdot\widetilde{Q}$
		\STATE  SVD decomposition: $U,S,V=SVD(H)$
		\STATE Rotation matrix calculation: $R=V\cdot \mathbf{U}^\top$
		\STATE Reflection matrix detection
		\STATE Translation matrix calculation: $T=-R\cdot\mathbf{\bar P}^\top+\mathbf{\bar Q}^\top$
		\STATE Invert the robotic arm’s coordinate point set using a $P$ point set: $P1=R\cdot P+T$
		\STATE Error calculation: $\alpha=P-P_1$
		\IF{$\alpha=P-P_1<\beta$} 
		\STATE \textbf{return} $R$, $T$
		\ELSE 
		\REPEAT
		\STATE Reject the set of points with significant errors, and repeat steps 3 through 11
		\UNTIL {$\alpha=P-P_1<\beta$} 
		\ENDIF
		\STATE \textbf{return} $R$, $T$
	\end{algorithmic}  
\end{algorithm}

After the hand-eye calibration, a 3D reconstruction of CT images of the hip bones was performed (as shown in Figure 3). In the reconstructed 3D model, the abduction angle, forward inclination angle, and cartilage thickness of the acetabular grinding were determined to plan the grinding path of the end actuator of the robotic arm. The 3D reconstruction model was then converted into a 3D point cloud, with the vertices of each marker ball selected, and the contours of each marker ball in the image coordinate system segmented using the Kd-Tree algorithm. Through the least squares method, the contour of each marker ball in the image coordinate system was fitted, and its centre coordinates were obtained. At the same time, the actual hip model was placed in the field of view of the optical locator, and then the spherical coordinates of the marker balls were collected. The transformation relationship between the optical locator and the image coordinate system was then calculated using the principle of Algorithm 1 to complete the matching between the image and the actual hip model. 
\begin{figure*}
	\centering
	\includegraphics[width=17.5cm]{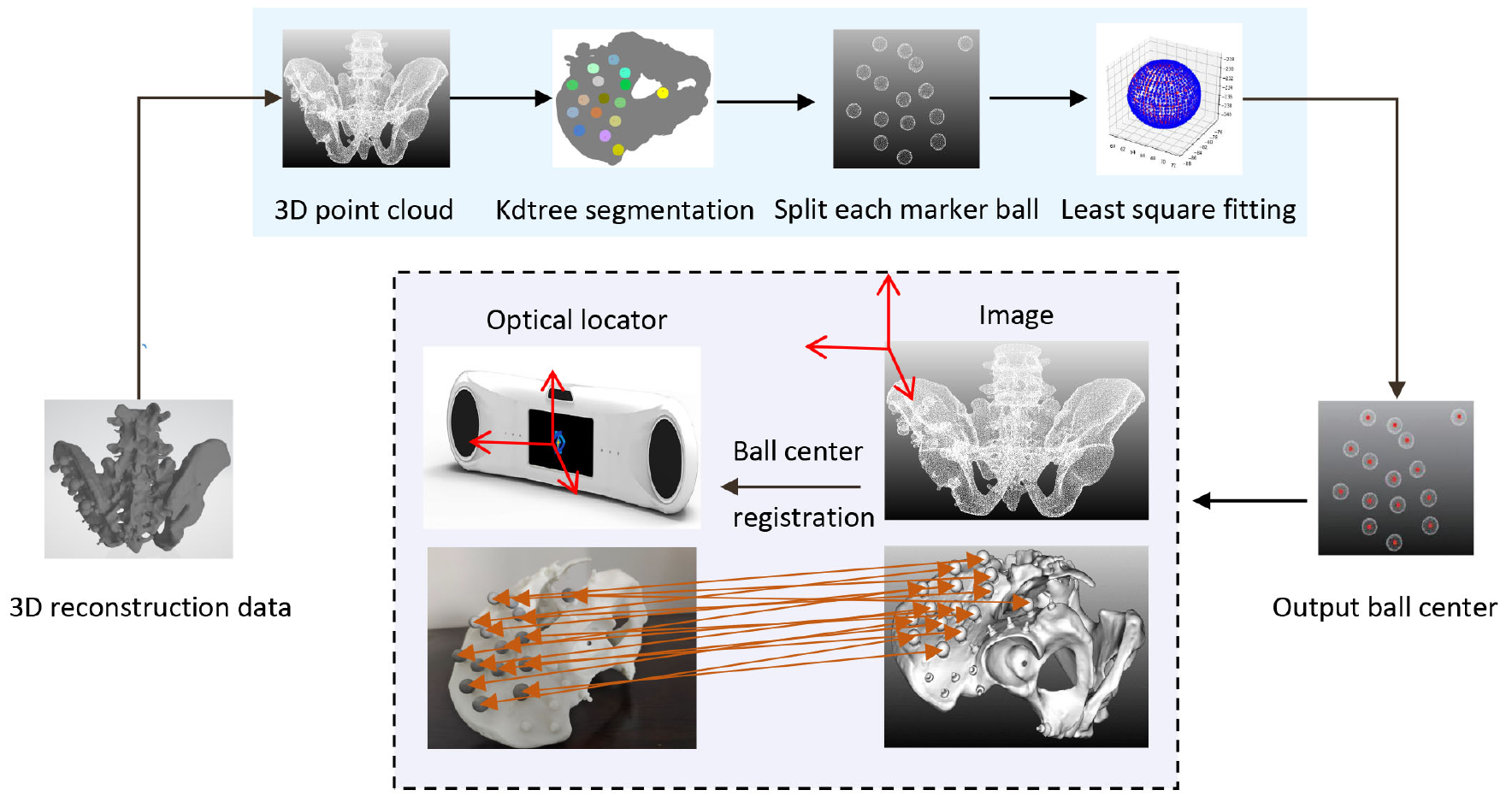}
	\caption{Surgical registration based on image registration.}
	\label{fig:3}
\end{figure*}
\subsection{End Actuators}
The end-effector of the surgical robot is used in the process of acetabular grinding. Its structural composition is shown in Figure 4.
\begin{figure}
	\centering
	\includegraphics[width=8.5cm]{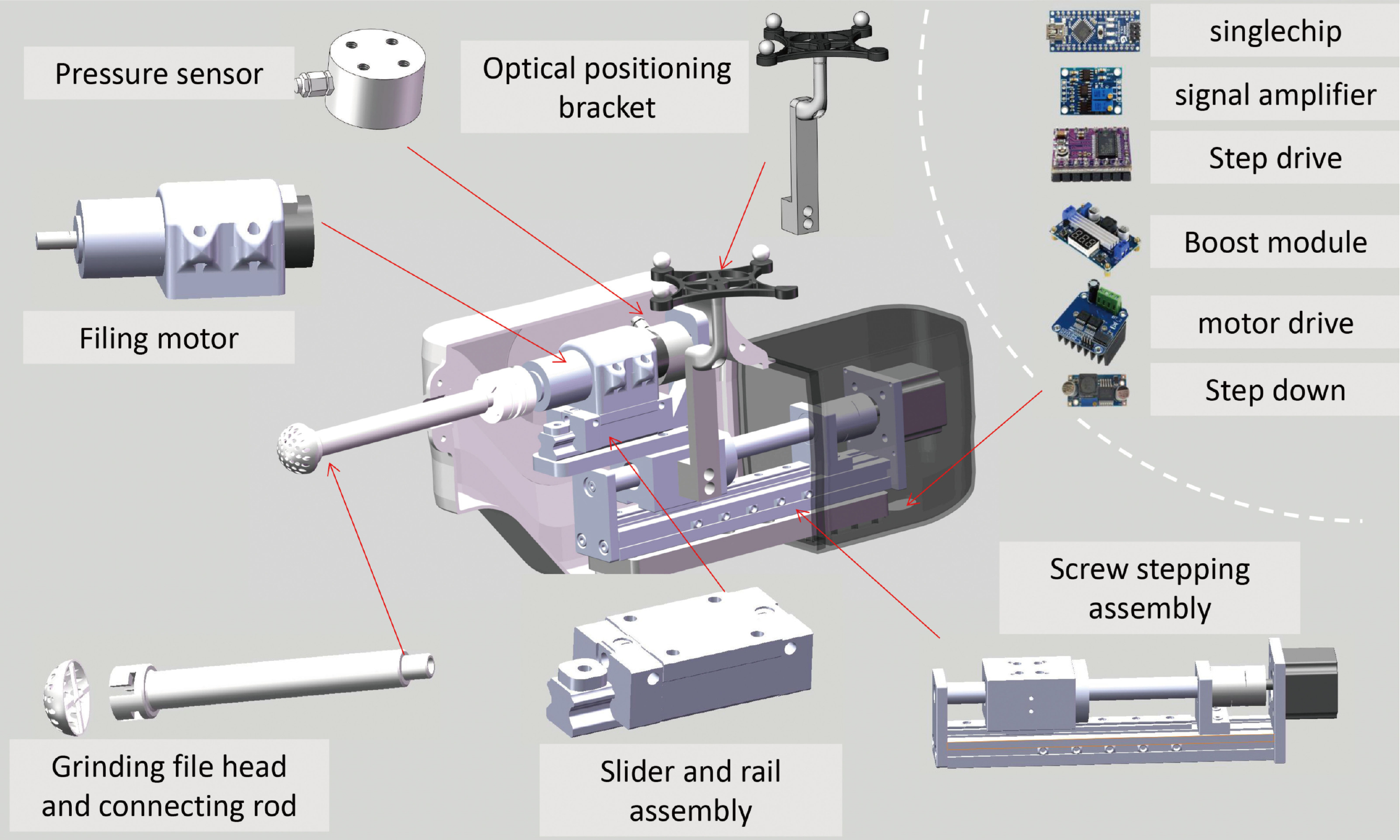}
	\caption{Structure composition diagram of the end-effector.}
	\label{fig:4}
\end{figure}
As shown in Figure 5, the user could obtain the movement information of the end-effector and contact pressure using the remote host computer and send control instructions to the microcontroller system through the wireless module. The microcontroller system sent control signals to both the stepper motor and the acetabulum reamer motor. During this process, we used optical positioning systems to track the marker balls, visualized their movements in real time, and passed them to the host computer controller. Simultaneously, a pressure sensor was used to obtain the contact pressure of the reamer as feedback. A database was then established to store the parameters corresponding to the reamer pressure and filing amount in a clinical experiment. If the pressure reached a threshold, the end-effector could be stopped in time to avoid excessive pressure during the grinding process, which would otherwise harm the patient. In future studies, with more sufficient data collected and more advanced control algorithms developed, the acetabulum reamer will grind more safely. 

\begin{figure}
	\centering
	\includegraphics[width=8.5cm]{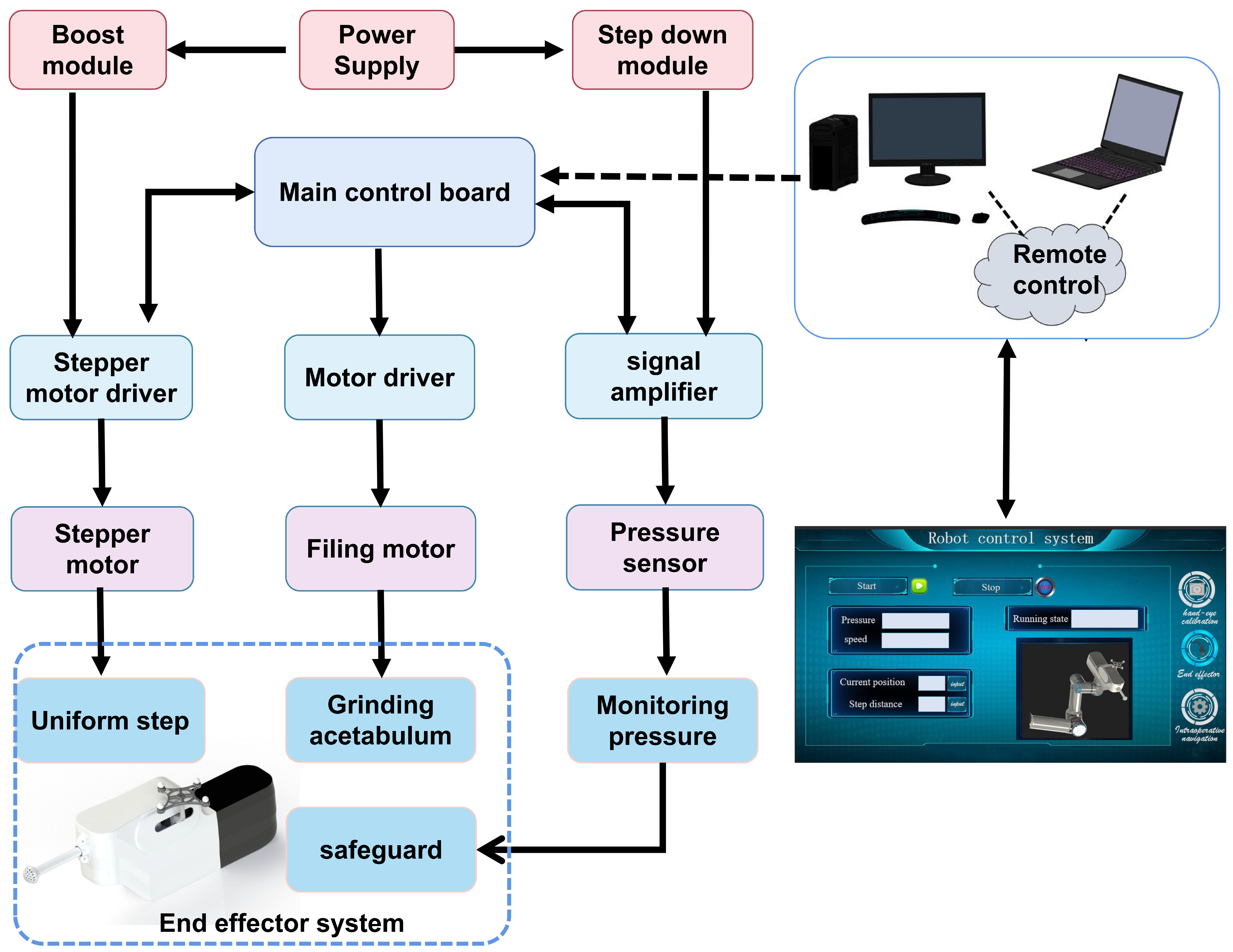}
	\caption{Principle of end-effector.}
	\label{fig:5}
\end{figure}

\subsection{Intraoperative Measurement and Navigation}
This study proposes a control scheme for the spatial measurement of key points in hip replacement surgery. The hip model was placed in the lateral recumbent position. In the self-designed position monitoring calibration tool, the established human body coordinate system of the patient was used as the reference coordinate system, enabling real-time changes in the intraoperative patient position to be monitored using the display of the surgical navigation interface. A rigid body coordinate system was established using three noncollinear marker balls at the end of the device. The end-effector was parallel to the desktop and placed according to the optical positioning system. The spatial coordinate information was initialised to zero. When the reamer on the end-effector was positioned, the rigid body coordinate system was visualised in real-time relative to the spatial information. This could assist surgeons to determine whether the grinding angle is within the normal range. To improve the accuracy of the grinding angle measurement, we proposed a method for adjusting the robotic arm attitude based on real-time optical feedback. After the transformation relationship between the optical locator and the base coordinate system of the robotic arm is obtained according to Algorithm 1, the desired value of the angle of the robotic arm could be calculated via the input of the desired value into the optical locator coordinate system. The error between the desired and true values, as obtained using the optical locator, could be added to the desired value. Thus, the angle of the robot arm was adjusted based on the error as the input value. This procedure was iterated continuously until the error satisfies the requirements.

Subsequently, an intraoperative image-free measurement scheme based on optical positioning probes was proposed. After the length of the femoral neck was measured using a probe during the operation, an appropriate femoral ball prosthesis model could be selected in real-time to effectively reduce the postoperative leg length discrepancy. Then it was further verified according to the difference in the length of both lower limbs.

\section{METHODOLOGY}
The research protocol for the complete surgical procedure was as follows: Preoperative - A hip model with optical spheres (Figure 6b, male, PVC material) was prepared, and it was a 3D reconstruction based on a preoperative CT scan created to determine the planned surgical path. Intraoperative - A position monitoring tool was first installed to complete the calibration process. Clamps were used to secure the hip bone to the operating table in the lateral recumbent position (Figure 6a), and the robot completed the calibration positioning. The intraoperative precision navigation system measured the grinding angle in real-time (e.g. abduction angle 40°±10°, forward inclination angle 15°±10°), thus verifying the control function of the end-effector during the grinding process. 
For this study, we used the intraoperative navigation system on a 1:1 skeletal model of the human body (Figure 6e, male, PVC material). Protocol: The optical probe (Figure 6d) was used to measure the length of the femoral neck, lengths of both lower limbs, as well as other parameters. The caliper and combined angle ruler were used to measure the length of the femoral neck for comparison. The ankle was fixed above the calibration pad. A piece of millimetre paper was placed under the pad (Figure 6f), and a fixed reference block was positioned in the hip (Figure 6g). Several different experimental situations were provided, e.g. the two lower limbs move at different distances from the pad, to test the accuracy of the measurement scheme.

\begin{figure}
	\centering
	\includegraphics[width=8.5cm]{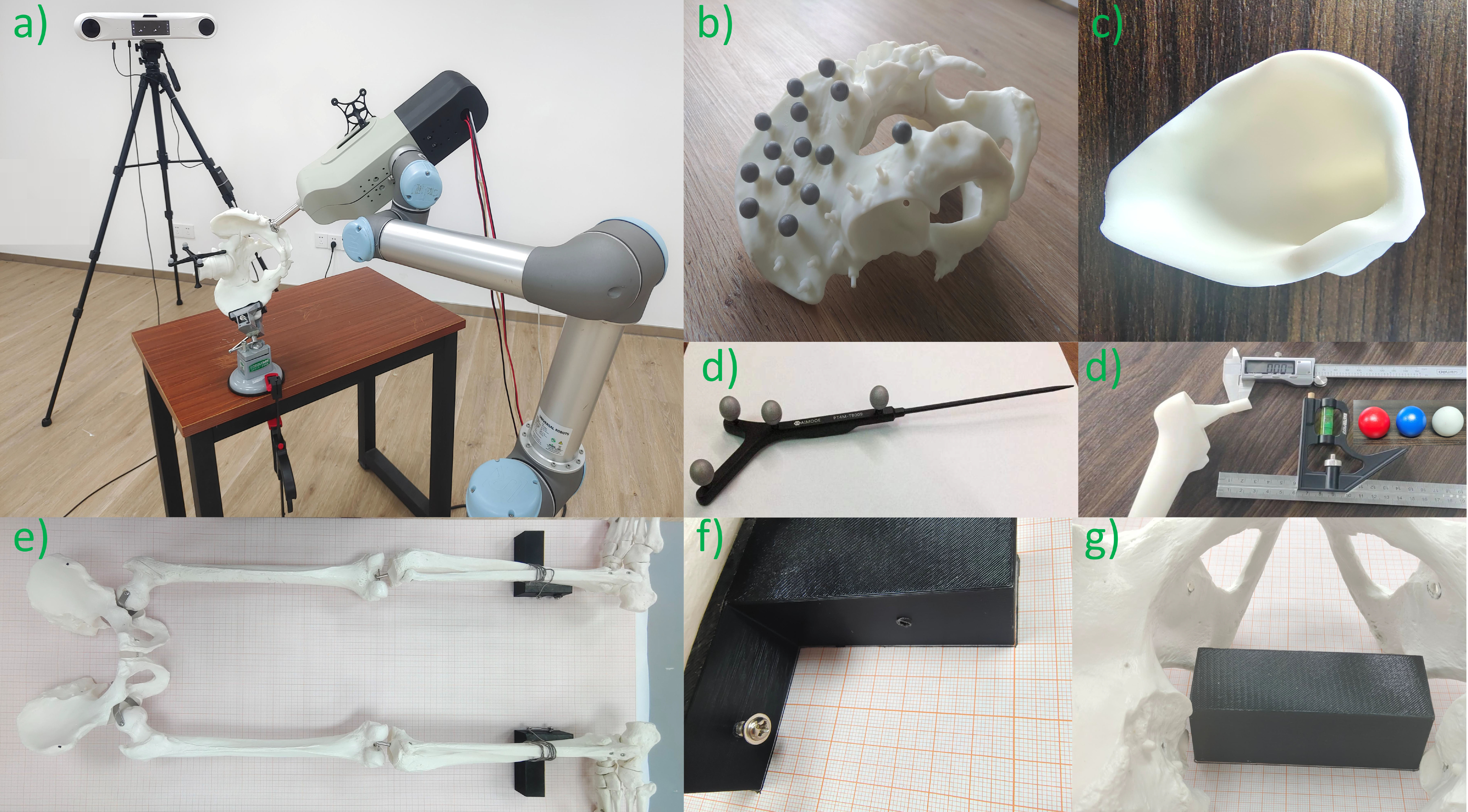}
	\caption{Simulation of experimental protocol for positioning and calibration system during surgery.}
	\label{fig:6}
\end{figure}

\subsection{Test 1}
The purpose of the first test was to evaluate the accuracy of the positioning of the robot. As shown in Figure 2, we used a self-developed calibration tool to verify the transformation accuracy of the entire machine based on the optical locator, robotic arm, and image space coordinate system. \\
\indent1) The operator selected the initial posture  $ A_{0}$ of the calibration tool at the end of the robotic arm before surgery \cite{zheng2020novel}, which was in the field of view of the optical locator. Once the initial posture was determined, the calibration tool was moved by a certain step length in subsequent calibration steps. \\
\indent2) Based on the error between the inversed and actual robot arm coordinates, the threshold value was set to 2 mm. The system then calculated the optimal rotation matrix $R$ and translation matrix $T$ between the point sets in the coordinate system of the manipulator and the optical positioner.
\subsection{Test 2}
The purpose of this study was to evaluate the functionality of the grinding angle measurement of the end-effector in real-time during the surgery. \\
\indent1) The hip was fixed onto the experimental platform, and the grinding and filing angles were measured in real-time during the simulated operation.

\indent2) After the surgeon planned the grinding angle in the image space, the angle was converted to the coordinate system of the robotic arm by the optical locator. The robot autonomously reached the target position accordingly, and the optical locator read the spatial position of the optical sphere at the end of the device in real-time, thereby obtaining the angle error value. The desired value of the optical locator, pulsing the error value, was used as the input, with the angle error threshold value set to 0.5°. Iteration continued until the experimental requirements were satisfied. 

\subsection{Test 3}
This test was performed to verify the control functionality of the end-effector. After the arm was positioned in the right place, it was locked. The reamer was then moved by the stepper motor. When it contacted the acetabular, the grinding motor automatically starts. Driven by the stepper motor, it grounded the hip bone evenly in a fixed direction. Simultaneously, real-time pressure detection was performed, with the safety threshold set to 30~N. When the pressure reached the threshold value, the grinding process stopped and the system was reset.
\subsection{Test 4}
This test was performed to evaluate the intraoperative measurement of femoral neck length and differences in lower limbs. The experimental setup is shown in Figure 7. First, femoral neck length was measured using an optical positioning probe. The measurements were repeated 10 times. Second, a scenario was given simulating the measurement of the difference between the two legs after surgery, in which the left and right legs were moved along the longitudinal direction of the millimetre paper by 5 mm and 10 mm, alternately. This step was repeated 10 times. \\
\indent1) As shown in Figure 7a, a 3D printed model (composed of a femur stalk inserted into a femur) was used to simulate anatomy measurement during surgery. The actual length $c$ of the rotation centre of the femoral head to the femoral surface was obtained using the end of an optical probe, which measured the distance from the highest point to the osteotomy surface, subtracted the radius of the femoral head prosthesis (Figure 7b). The coordinates $P_1$ ($x_1$, $y_1$, $z_1$), $P_2$ ($x_2$, $y_2$, $z_2$), and $P_3$ ($x_3$, $y_3$, $z_3$) of the three reference points were obtained sequentially from the osteotomy surface of the affected limb, and the coefficients $A$, $B$, $C$, and $D$ of the osteologic surface equation were calculated using equation (1).
\begin{align}
Ax+By+Cz+D=0
\end{align}
\indent The measurement point of the femoral spheroid prosthesis was the highest point, $P_4$ ($x_4$, $y_4$, $z_4$), on the femoral spherical prosthesis. The formula (2) was used to calculate the distance between the osteotomy surface of the affected limb and the measurement point $e$.
\begin{align}
e=\frac{Ax+By+Cz+D}{\sqrt{A^{2}+B^{2}+C^{2}}}
\end{align}
where $x$, $y$, and $z$ represented the coordinate values of the measurement points; and $e$ represented the distance between the osteotomy surface of the affected limb and the measurement point.
\begin{figure}
	\centering
	\includegraphics[width=8.5cm]{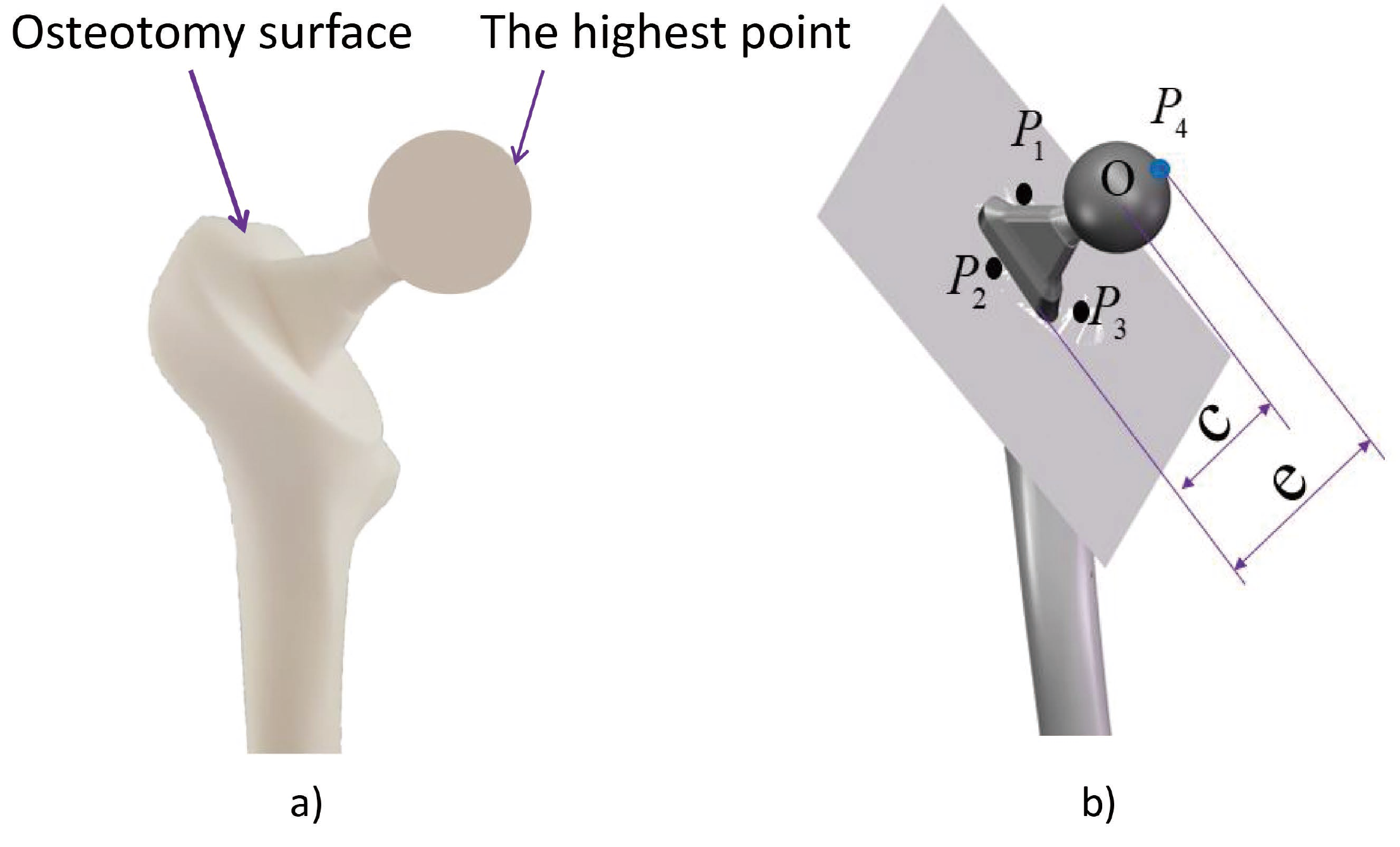}
	\caption{Intraoperative measurement of femoral neck length.}
	\label{fig:7}
\end{figure}

2) The simulation compensated for errors in the operation by selecting different types of femoral head prostheses.

3) After the prosthesis was installed, the length of the affected leg was measured in real time. We compared the values with those obtained from preoperative measurements.  

\section{RESULTS \& DISCUSSION}

\subsection{Calibration of Positioning Control by Surgical Robot}

Preoperative surgical path planning (Figure 8) was performed to determine the optimal grinding angle of the robot, with the safe point, grinding point, and target point being shown in the image space.
\begin{figure}
	\centering
	\includegraphics[width=8.5cm]{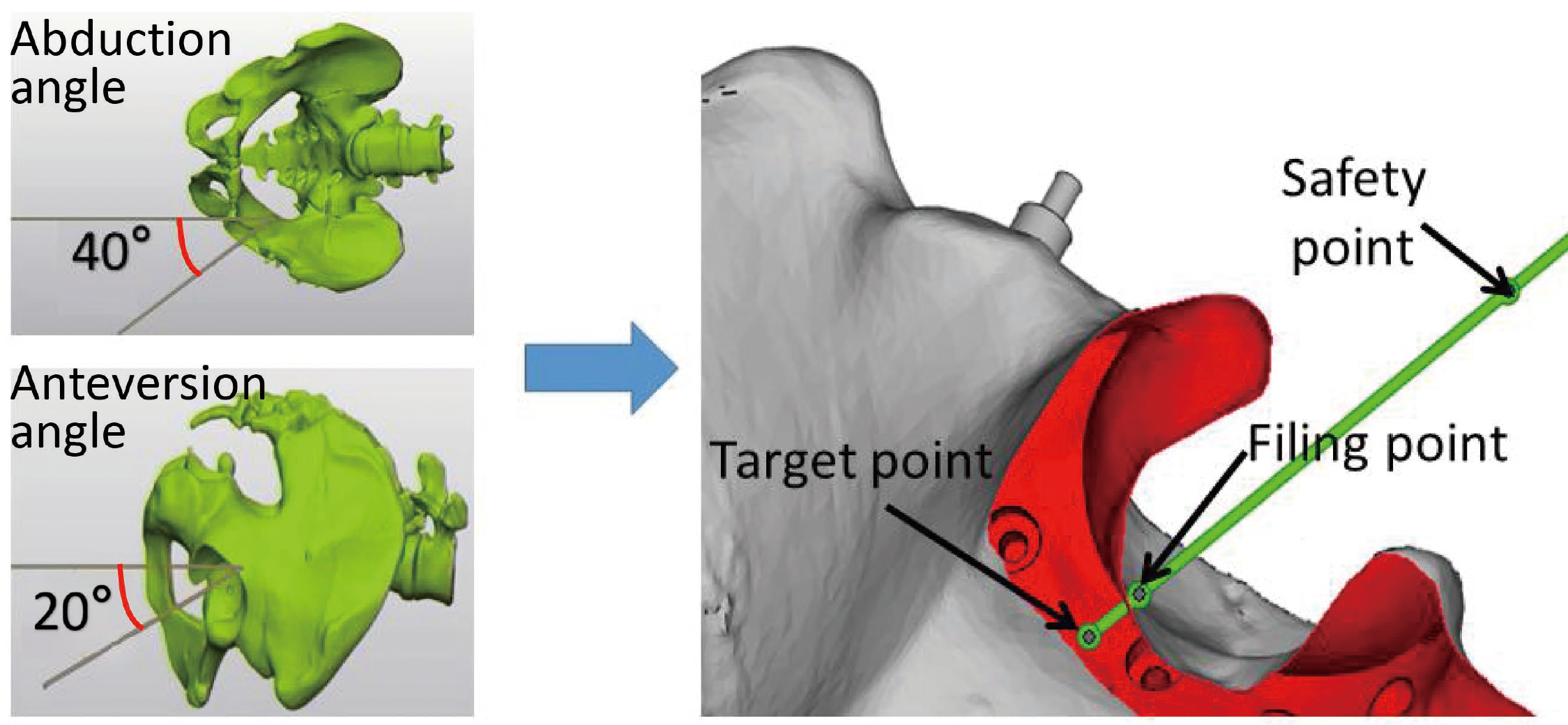}
	\caption{Preoperative robot reamer entry path planning.}
	\label{fig:8}
\end{figure}
\begin{figure}
	\centering
	\includegraphics[width=8.5cm]{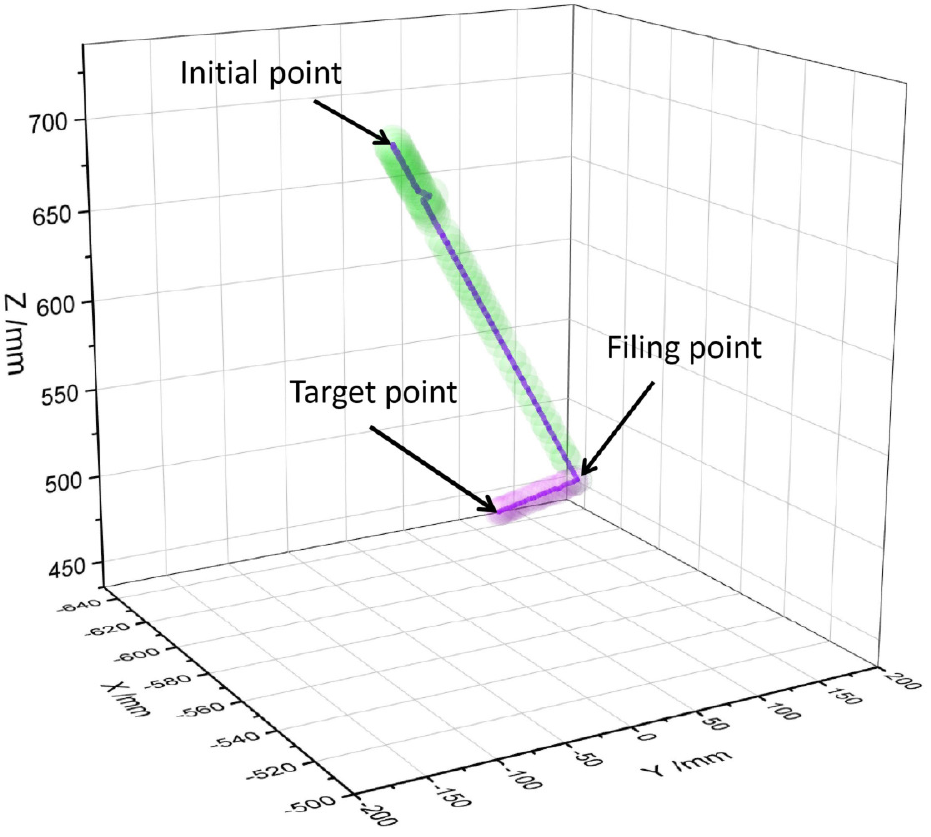}
	\caption{Trajectory of robotic arm.}
	\label{fig:9}
\end{figure}

The actual motion trajectory of the reamer at the end of the manipulator during surgery (Figure 9) required the calculation of the spatial distance between its desired and measured positioning values. As shown in Figure 10, the average Euclidean errors generated in the experiment, included 0.72~mm in the x-axis direction, 0.71~mm in the y-axis direction, and 0.69~mm in the z-axis direction, was 1.51~mm.

\begin{figure}
	\centering
	\includegraphics[width=8.5cm]{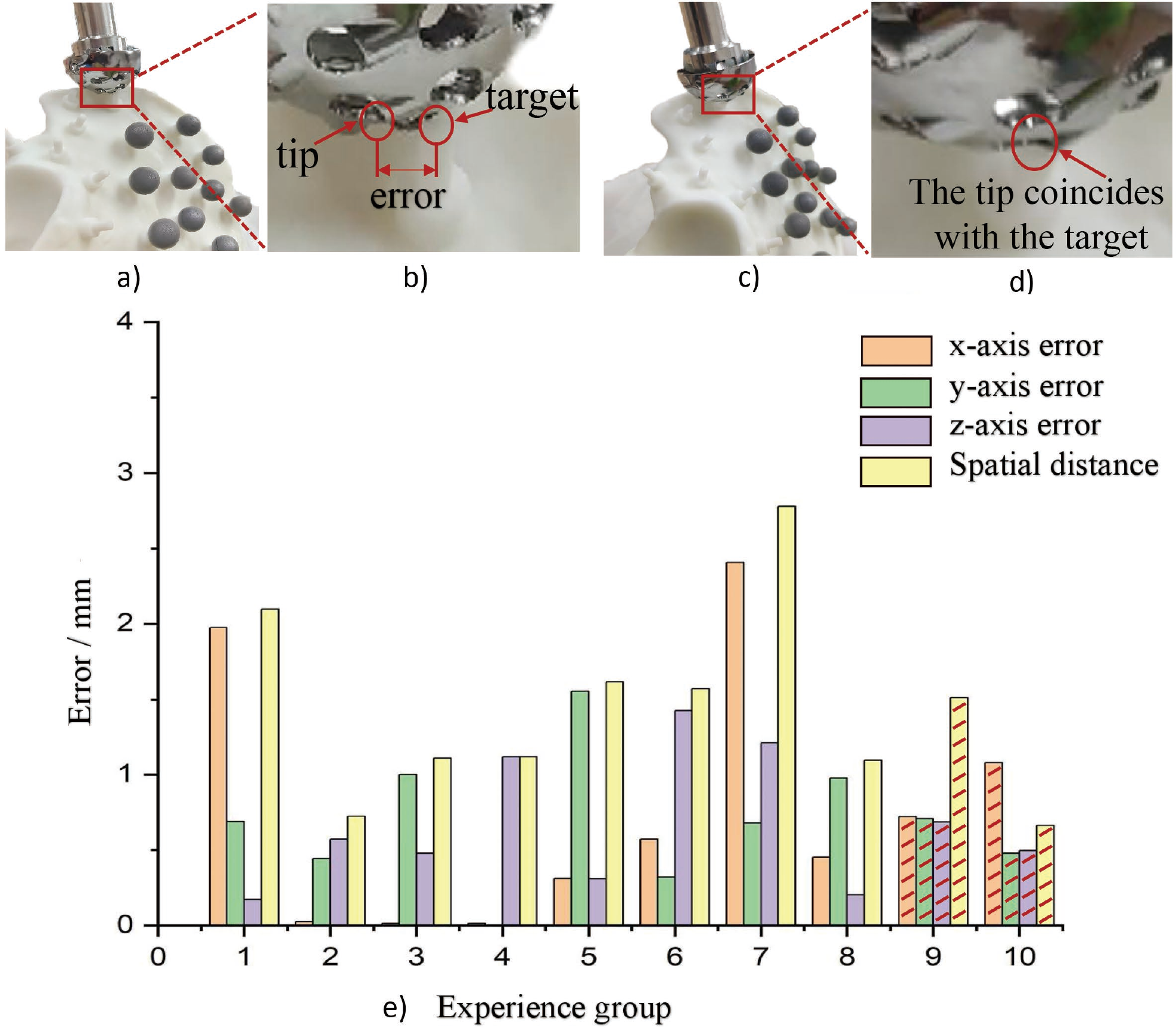}
	\caption{Surgical robot calibration experiment. a) Expected value of robot arm positioning.  b) local enlarged view of a). c) real value of robot arm positioning. d) local enlarged view of c). e) histogram of positioning accuracy of grinding file tip. Zones 1 through 8 represent Euclidean distance directional error, and zones 9 and zone 10 represent mean error and standard deviation (SD).}
	\label{fig:10}
\end{figure}

\subsection{Evaluation of Robot Measurement Navigation System}
The measurement results of the surgical robot for the hip morphological parameters of the patient were as follows: When the robot reached the desired coordinate point, if no feedback adjustment was added, the average errors of the forward tilt angle and abduction angle (Figure 11a, Figure 11c) were 0.99° and 1.79°, respectively. After feedback was added to adjust the attitude of the robot, the average errors of the forward tilt angle and abduction angle (Figure 11b, Figure 11d) were 0.22° and 0.15°, respectively. After the robot arm reached the desired value, the output data of the optical positioner was fed back to improve the accuracy of the grinding angle. According to CAD models of the femoral neck prostheses, the sizes of the red, blue, and white femoral heads were 58, 59, and 60 mm, respectively. As shown in Figure 12, the average error of the distance from the highest point of femoral bulb prosthesis to the osteotomy surface, as measured using optical probes, were 0.071, 0.118, and 0.087~mm, respectively. For comparison, the average errors measured using the mechanical equipment were 0.287, 0.278, and 0.307~mm. Thus, it was demonstrated that the optical probe had much higher precision and could effectively assist surgeons in performing THA surgery.

\begin{figure}
	\centering
	\includegraphics[width=8.5cm]{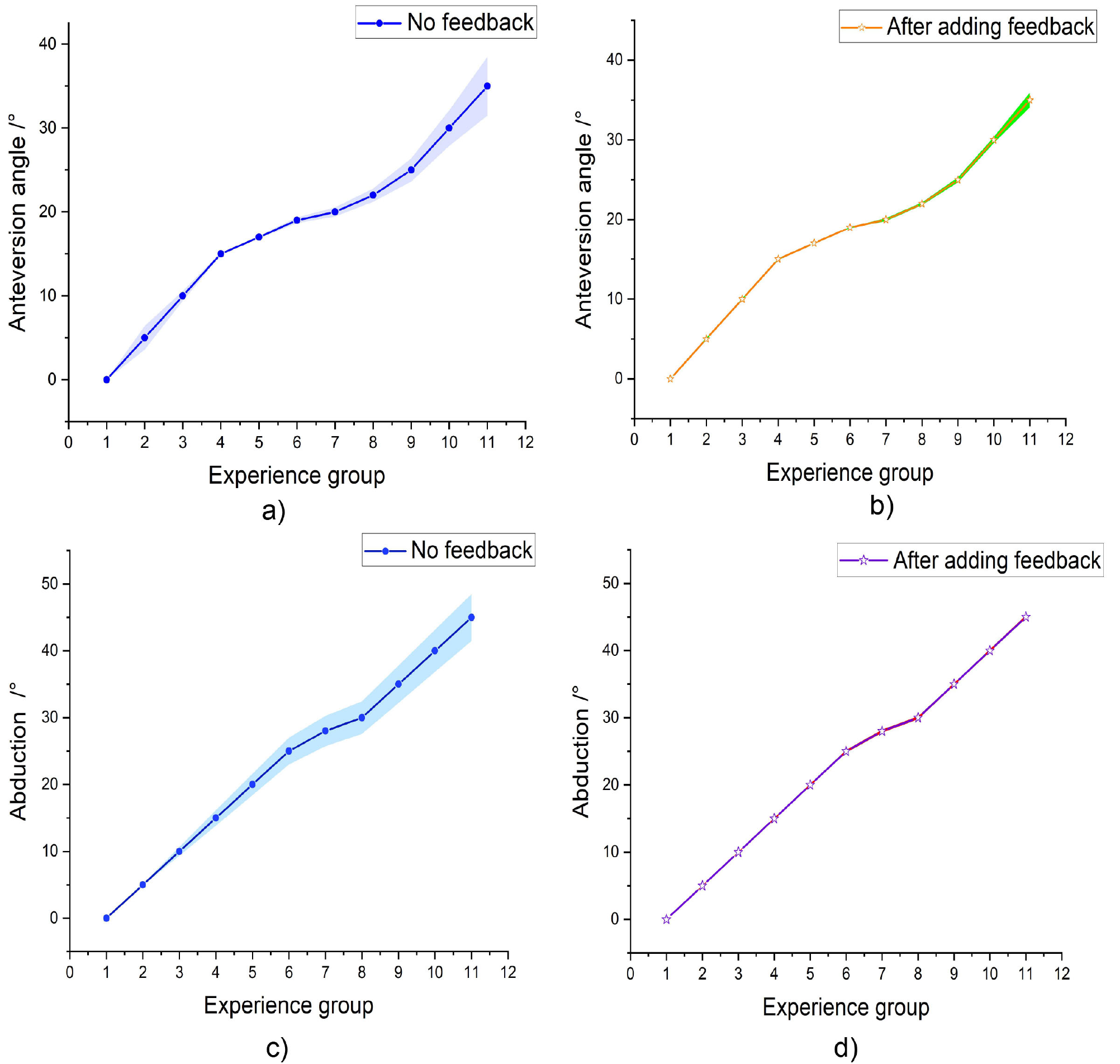}
	\caption{Real-time measurement of grinding angle after positioning of the robotic arm. a) and c) are error band diagrams of the anteversion and abduction angles without feedback. b) and d) are the error band diagram of the anteversion and abduction angles after feedback.}
	\label{fig:11}
\end{figure}

\begin{figure*}
	\centering
	\includegraphics[width=17.5cm]{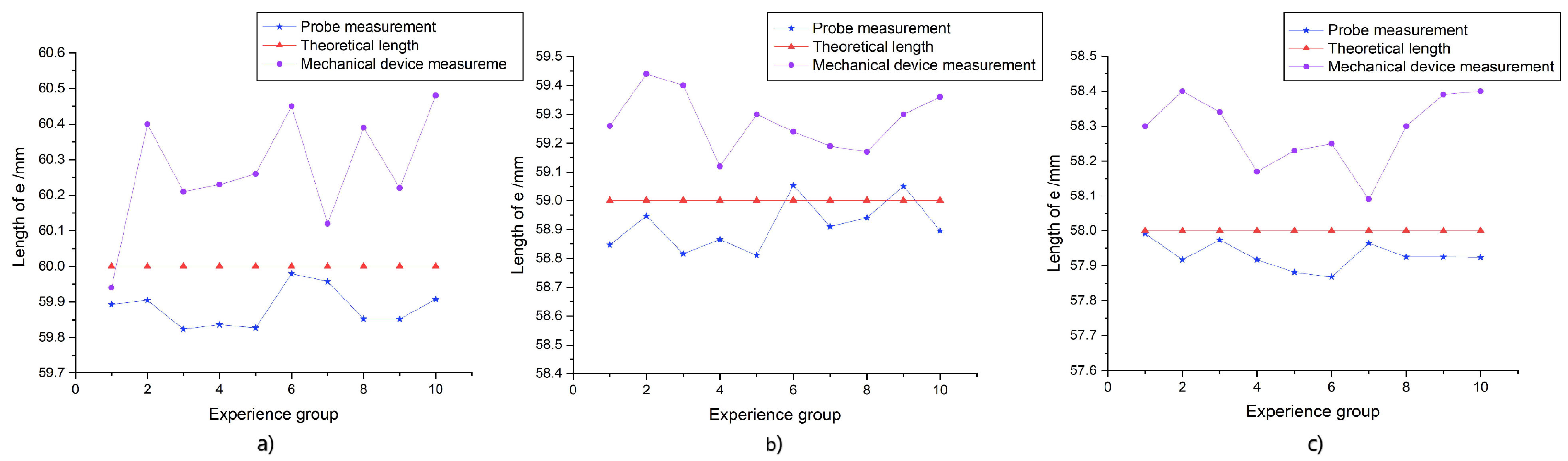}
	\caption{Comparison of distance e from highest point of femoral bulb prosthesis to osteotomy surface. a), b), and c)  indicate red, blue, and white femoral heads, respectively, corresponding to different measured values e.}
	\label{fig:12}
\end{figure*}

Based on data in Table 1, the following conclusions could be drawn. When the two lower limbs did not move, the average difference between the two lower limbs was 0.17~mm. When the affected limb moved down 5~mm and 10~mm longitudinally, the average differences between the two lower limbs were 4.891 and 10.299~mm, respectively. In the aforementioned experiment, the measurement error of the mean difference between the two lower limbs, after discounting the movements of the affected limb, was within 0.3~mm (0.219 and 0.299mm respectively).

\begin{table}[h]	
	\caption{Measurement of both lower limbs}
	\label{tab7} %设置表的引用标签
	\begin{tabular}{ccccc}
		\toprule[1.5pt]
		%%%%%%%%%%%%%%%%%%%%%%%%%%%%%%%%%%%%%%%%%%%%%
		%\textwidth 是每一行的宽度.[0.1\textwidth]设定单元格宽度
		% [c]  单元格文本居中对齐
		% {name} 单元格内容
		%%%%%%%%%%%%%%%%%%%%%%%%%%%%%%%%%%%%%%%%%%%%%
		\makebox[0.008\textwidth][c]{Name} & \makebox[0.008\textwidth][c]{Variation} & \makebox[0.018\textwidth][c]{Group}
		& \makecell{Average length }& \makebox[0.007\textwidth][c]{SD}             \\
		\midrule[1pt]
		Left limb  & No & 10& 832.128 mm & 0.105 \\
		Right limb  & No & 10& 832.298 mm & 0.083 \\
		Left limb  & Down 5 mm & 10& 837.189 mm & 0.024  \\
		Left limb  & Down 10 mm & 10& 842.597 mm & 0.038 \\
		\bottomrule[1.5pt]
	\end{tabular}
\end{table}

\subsection{End-Effector Control}
The motor of the end-effector implied the grinding scheme according to the instructions for grinding the model of cartilage and hipbone. As shown in Figure 13a and 13c, the end-effector file could fulfill normal ground. As shown in Figure 13b and 13d, it could stop and exit with an emergency shutdown when reaching a pressure value of 30 N.

\begin{figure}
	\centering
	\includegraphics[width=8.5cm]{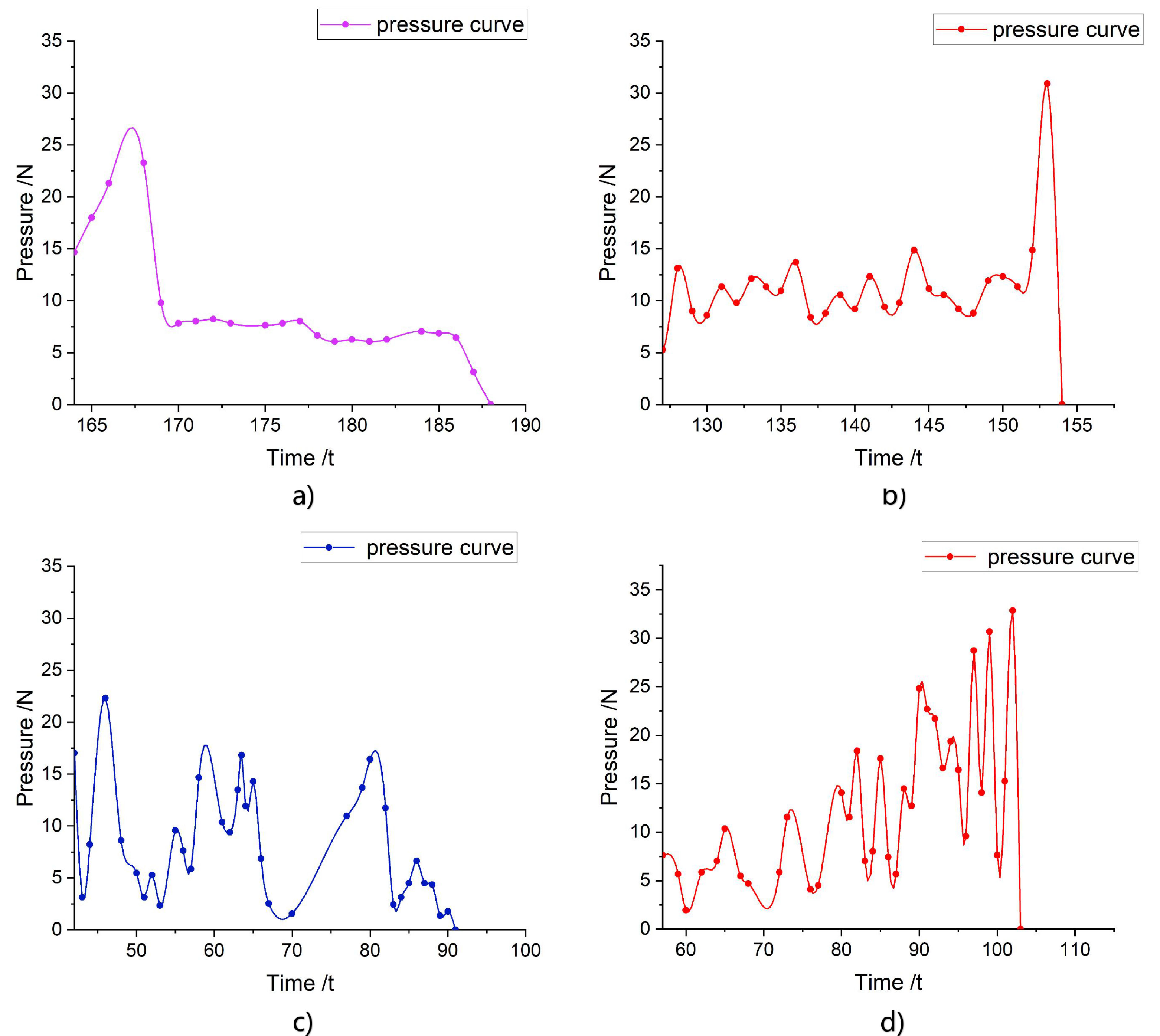}
	\caption{Grinding pressure change curve. a) normal pressure withdrawal during grinding of cartilage model; b) excessive pressure withdrawal during grinding of cartilage model; c) normal pressure withdrawal during grinding of bone model; d) excessive pressure withdrawal during grinding of bone model.}
	\label{fig:13}
\end{figure}

\section{CONCLUSION}

This paper proposes a robotic system for THA surgery equipped with functionalities for high-accuracy positioning of the preoperatively planned target and grinding angle during surgery. In addition, we propose an accurate intraoperative measurement system based on optical probes, which measures the femoral neck length and the difference between the two lower limbs in real-time during the operation. This can effectively assist surgeons in making surgical decisions and selecting appropriate prostheses to minimise preoperative errors. A limitation of this study is that the results are only achieved by simulated experiments. In the next step, we will apply the developed system to clinical trials for further verification. 

%\begin{sloppypar}
%\section{Funding}
%\end{sloppypar}
%% Loading bibliography style file
%\bibliographystyle{model1-num-names}
%\bibliographystyle{cas-model2-names}

\bibliography{cas-refs}
\clearpage

\section{APPENDIX}
\subsection{Desired femoral neck length}
As shown in Figure 14, the patient's right acetabulum was normal, but with defected in the left acetabulum development. According to the principle of symmetry, the position of the left osteotomy surface could be planned with reference to the right side before operation. If we measured the vertical distance e from the highest point of the femoral head to the osteotomy surface, we could then obtain the desired value $c_{0}$ of the preoperative femoral neck length from the formula $c_{0}=e_{0}-r$, where r was the rotation radius of the femoral head. Finally, according to the length of the femoral neck, the inner diameter of the femoral medullary cavity and the insertable depth of the femoral medullary cavity, the model of the prosthesis abled to be fit into the medullary cavity could be obtained from the manual. Figure 14 showed the template of the acetabular prosthesis implanted on the left side.
\begin{figure}
	\centering
	\includegraphics[width=8.5cm]{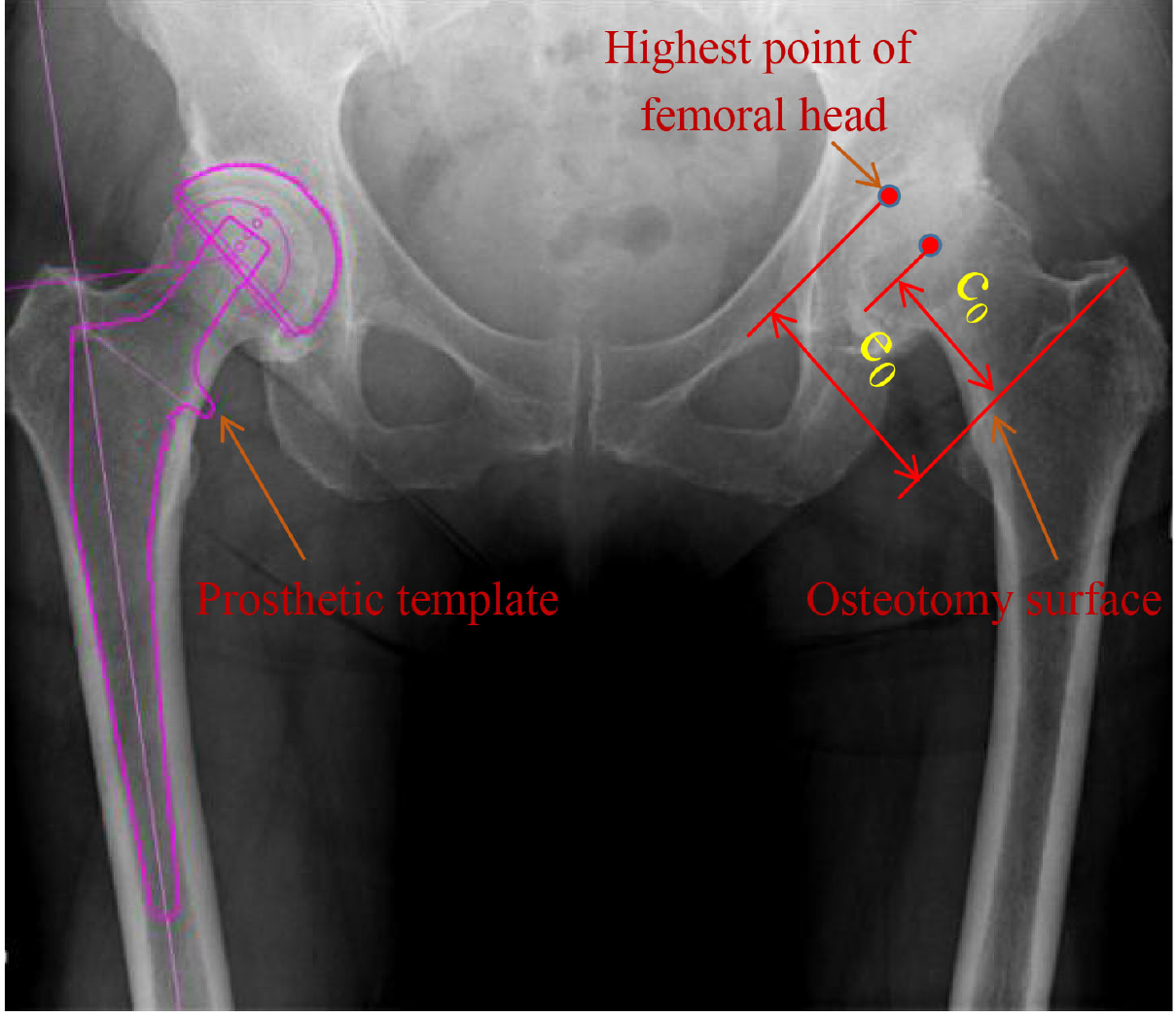}
	\caption{Preoperative planning of desired femoral neck length.}
	\label{fig:14}
\end{figure}

\subsection{Intraoperative Navigation Measurement}
As shown in Figure 15, after the femoral stem prosthesis was installed, the length $e$ from the highest point to the osteotomy surface was measured using a probe. Based on the desired value from preoperative measurement, a comparison between the selected femoral bulb and femoral neck prosthesis was performed. Thus, the final femoral bulb prosthesis could be accurately selected. We prepared three different depth models of red, blue, and white femoral head prostheses (Figure 16) and made reasonable adjustments to compensate for preoperative errors according to the femoral neck length measured in real time. Subsequently, the length of both lower limbs for the selected prosthesis was measured for further verification and feedback.
\begin{figure}
	\centering
	\includegraphics[width=8.5cm]{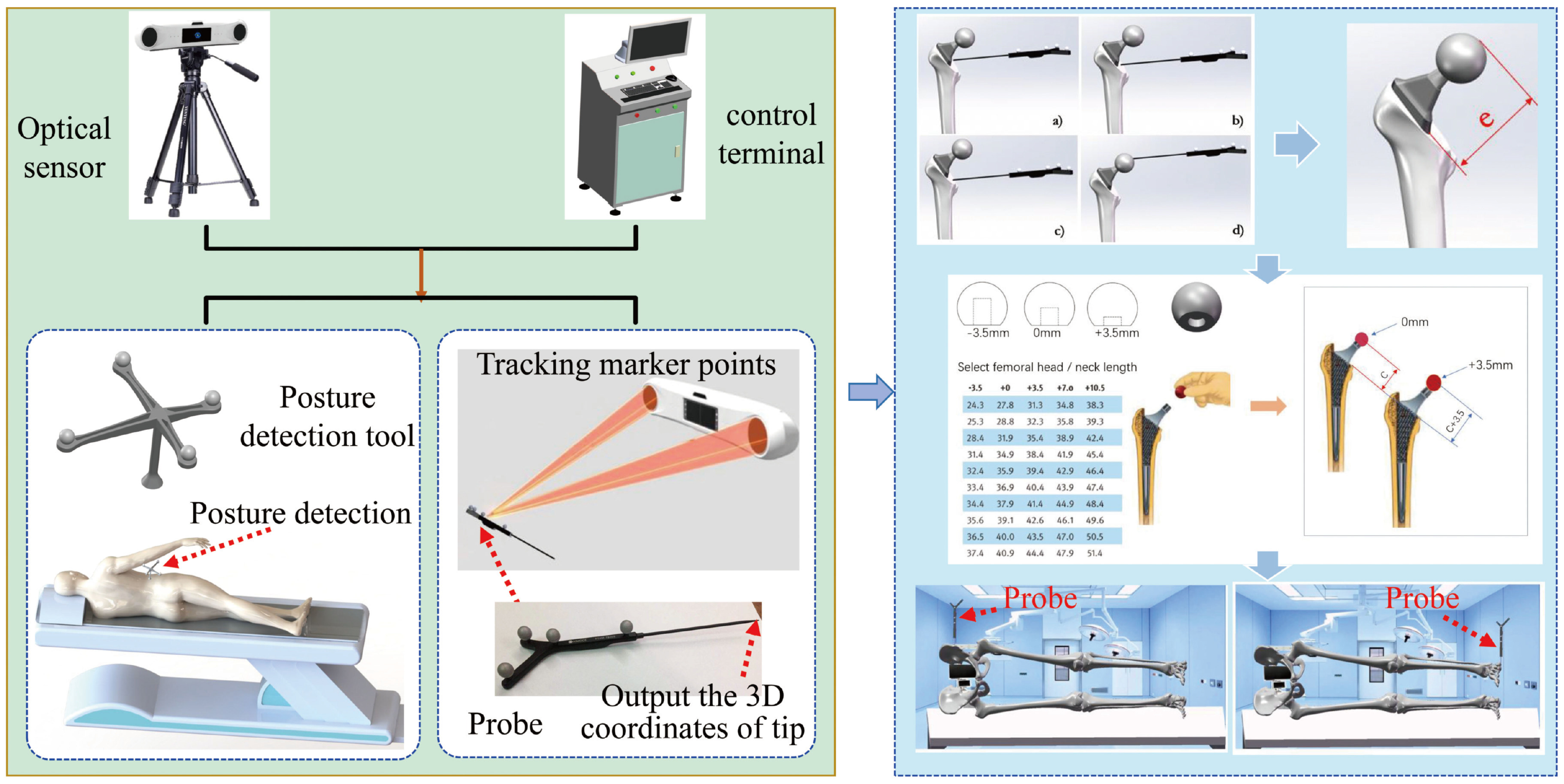}
	\caption{Intraoperative measurements and navigation based on optical positioning.}
	\label{fig:15}
\end{figure}
\begin{figure}
	\centering
	\includegraphics[width=8.5cm]{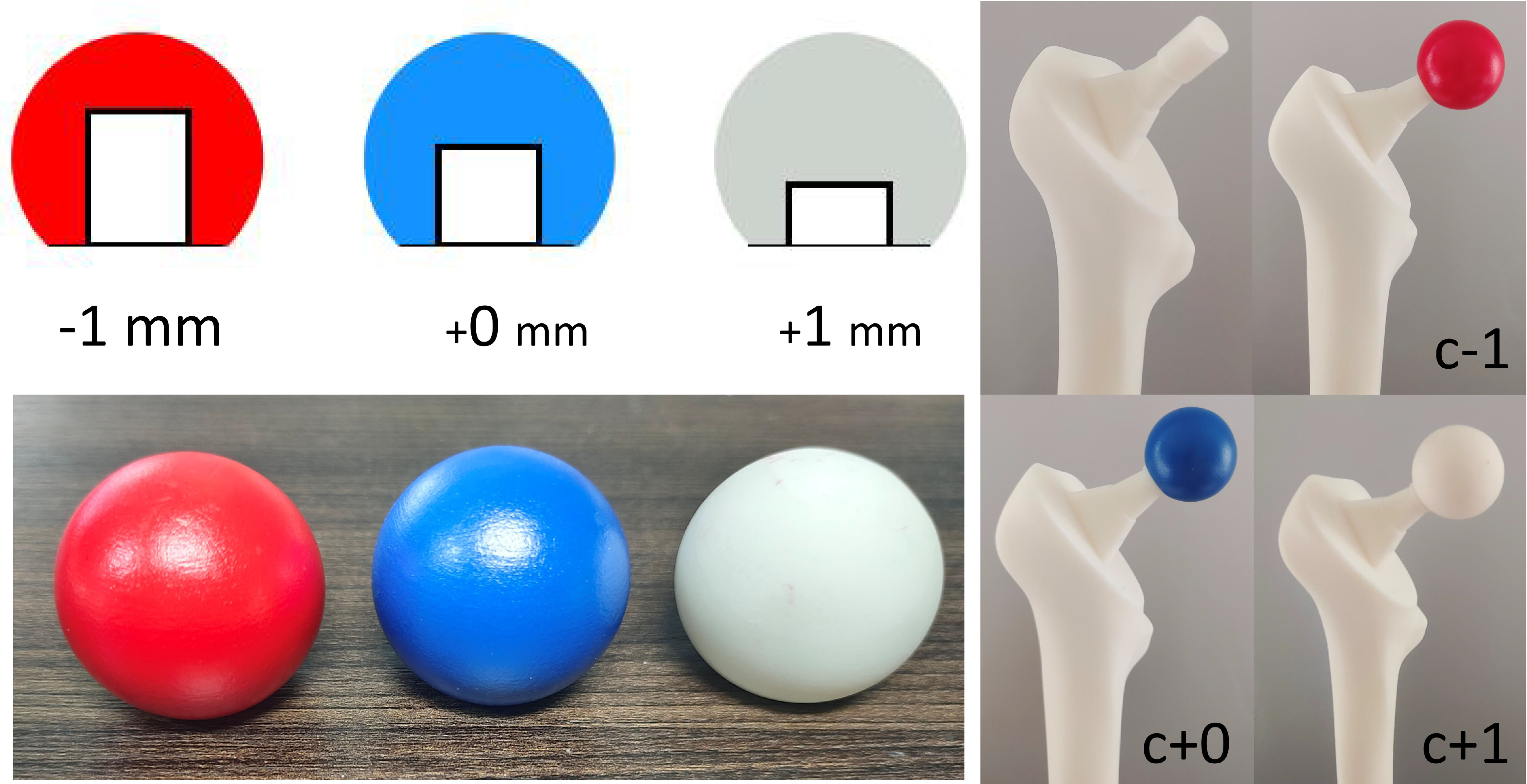}
	\caption{Selection of prosthetic specifications; $c$ is actual length from rotation  centre to osteotomy surface.}
	\label{fig:16}
\end{figure}

\subsection{Real-time Measurement of Grinding Angle}
The hip was fixed in the lateral recumbent position on the platform forceps, and calibration initialisation was performed on the end-effector parallel to the hip model (Figure 17a). The relative angles of the tools on the end-effector with attached marker balls were the abduction angle and forward tilt angle of the intraoperative reamer (Figure 17b). The placement of optical monitoring tools simulated intraoperative real-time observations of changes in hip position.
\begin{figure}
	\centering
	\includegraphics[width=8.5cm]{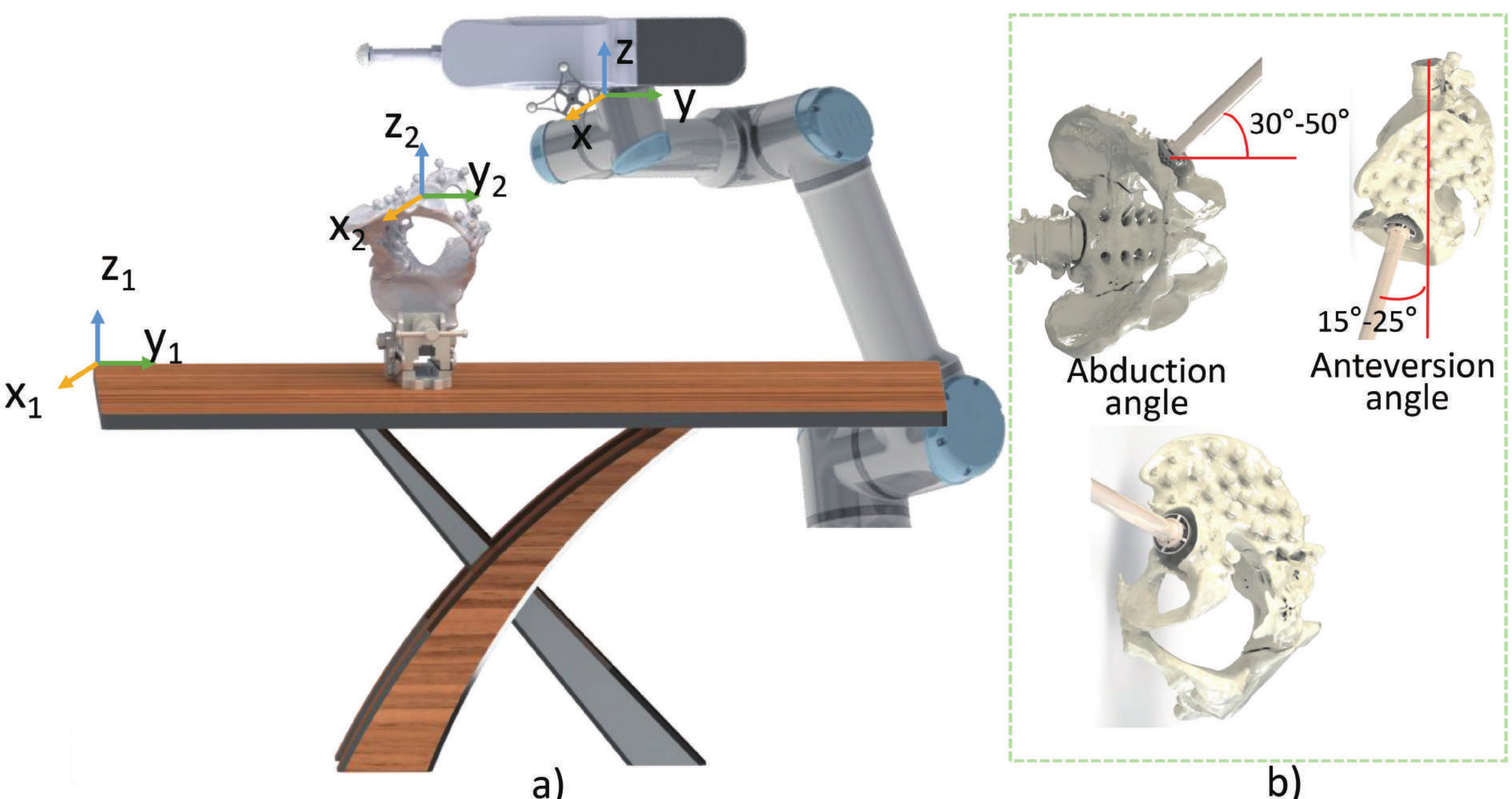}
	\caption{Simulation of calibration protocol for intraoperative localisation initialisation.}
	\label{fig:17}
\end{figure}

%\vskip3pt

%\bio{}
%Author biography without author photo.

%\endbio

\end{document}